\documentclass[conference]{IEEEtran}

\usepackage{cite}
\usepackage{amsmath,amssymb,amsfonts}
\usepackage{booktabs}
\usepackage{diagbox}
\usepackage{graphicx}
\usepackage{hhline}
\usepackage{multirow}
\usepackage{textcomp}
\usepackage[normalem]{ulem}
\usepackage[table,xcdraw]{xcolor}
\usepackage{url}

\useunder{\uline}{\ul}{}

\def\BibTeX{{\rm B\kern-.05em{\sc i\kern-.025em b}\kern-.08em
    T\kern-.1667em\lower.7ex\hbox{E}\kern-.125emX}}

\newif\ificdmanonymous
\icdmanonymousfalse
\begin{document}

\title{Perseus: Interactive Time Series Segmentation with Sparse Supervision via Stateful Memory}

\ificdmanonymous
\author{\IEEEauthorblockN{Anonymous Authors}}
\else
\author{
\IEEEauthorblockN{Ching Chang\IEEEauthorrefmark{1}\IEEEauthorrefmark{3},
Ming-Chih Lo\IEEEauthorrefmark{1},
Chiao-Tung Chan\IEEEauthorrefmark{2},
Wen-Chih Peng\IEEEauthorrefmark{1}, and
Tien-Fu Chen\IEEEauthorrefmark{1}}
\IEEEauthorblockA{\IEEEauthorrefmark{1}\textit{Department of Computer Science},
\textit{National Yang Ming Chiao Tung University}, Hsinchu, Taiwan}
\IEEEauthorblockA{\IEEEauthorrefmark{2}\textit{Department of Electrical and Control Engineering},
\textit{National Yang Ming Chiao Tung University}, Hsinchu, Taiwan}
\IEEEauthorblockA{\IEEEauthorrefmark{3}Corresponding author: blacksnail789521@gmail.com}
}
\fi

\maketitle

\begin{abstract}
Real-world systems, ranging from industrial manufacturing to wearable healthcare, generate multivariate time series with hierarchical states ranging from coarse regimes to fine-grained events.
Unlike zero- or few-shot segmentation, our setting uses dense state labels for model training.
Sparse expert prompts provide inference-time corrections that resolve sequence-specific ambiguities without retraining.
In practice, this feedback is grouped around selected events or transitions, leaving large portions of the timeline unprompted.
The prompt-based sliding-window baselines evaluated here are \textit{stateless} with respect to user interaction history: they use guidance only within the current window and cannot retain it across these gaps.
To address this, we propose \textbf{Perseus} (\textbf{Per}sistent \textbf{Se}gmentation with \textbf{U}ser \textbf{S}upervision), a framework that transitions from synchronous processing to \textit{asynchronous state management}.
Perseus decouples supervision from inference via a distinct Write-Read architecture: grouped user cues are asynchronously encoded into a persistent memory bank (Write), which is then actively queried by the inference engine (Read) to service unprompted windows.
This mechanism bridges supervision gaps by conditioning predictions on a global history of interactions rather than solely on local inputs.
Extensive experiments on six datasets demonstrate that while evaluated stateless prompting baselines degrade significantly under grouped supervision, Perseus maintains robustness and achieves up to 85\% accuracy improvement in multi-granularity settings.
Code and preprocessing instructions are available at \url{https://github.com/blacksnail789521/Perseus}.
\end{abstract}


\section{Introduction}
\label{sec:introduction}

\begin{figure}[t]
    \centering
    \includegraphics[width=\columnwidth]{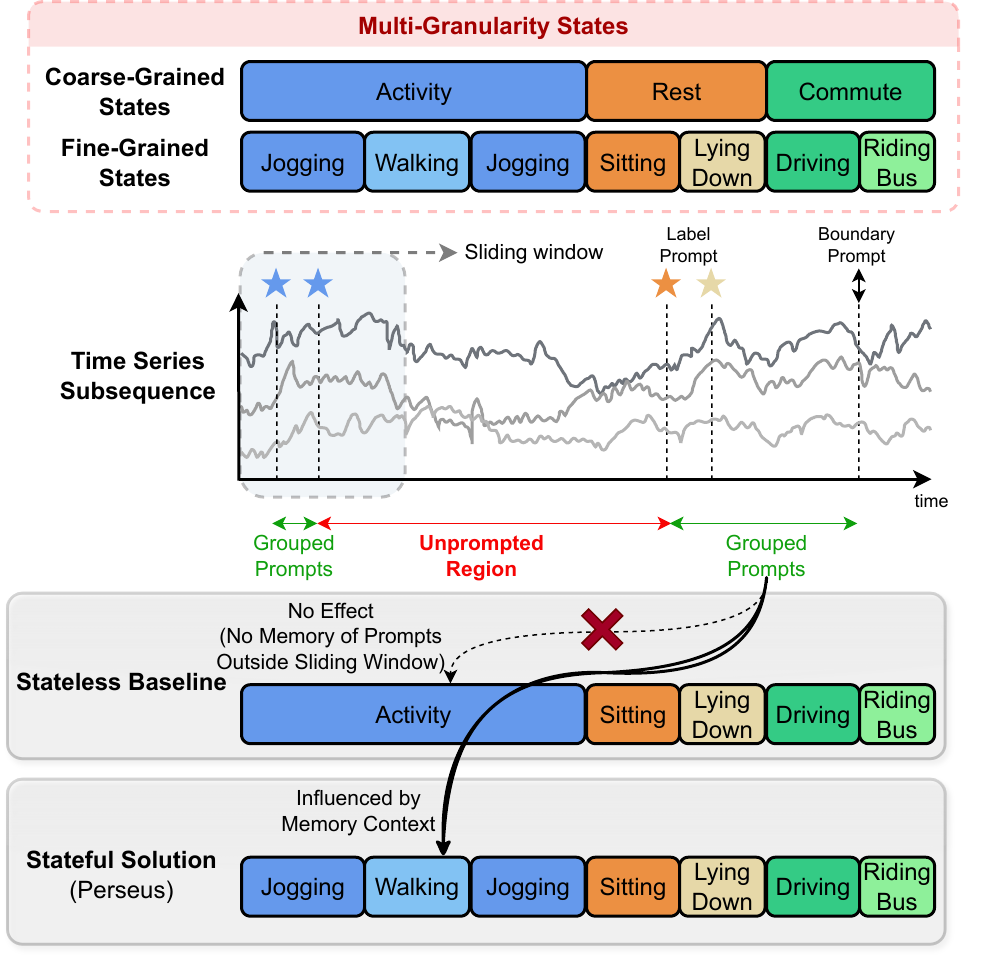}
    \caption{The unprompted region failure mode.
    Grouped supervision creates gaps where stateless prompting models cannot access prior interactions.
    Perseus uses stateful memory to retrieve that guidance in unprompted windows.}
    \label{fig:unprompt_region}
\end{figure}

Modern web platforms, mobile apps, and manufacturing systems generate rich multivariate time series capturing evolving user and device states.
Examples include smart home activity recognition \cite{smart_home_1, smart_home_2, timedrl_workshop}, financial market monitoring \cite{finance_1, finance_2, llm4ts_workshop}, and sports performance tracking \cite{sport_1, sport_2, timedrl}.
These states often exhibit multiple levels of granularity, from coarse regimes to fine-grained events \cite{wisdm, time2state, ticc, ts_reasoning_survey}.
For instance, long-term market cycles overlap with short-lived volatility spikes \cite{finance_1, text2freq}, while applications like anomaly detection rely on understanding these interactions \cite{coke, timeimm}.
To analyze such behaviors, domain experts increasingly use interactive tools to inject sparse feedback \cite{llm4ts, sam, run}.
However, unlike current model assumptions, real-world interaction is highly \textit{bursty and grouped}.
Practitioners rarely place prompts\footnote{In this work, we use the term "prompting" to refer to the injection of sparse signal cues (e.g., clicks, labels) to guide segmentation, distinct from natural language instructions used in Large Language Models (LLMs)~\cite{prompttss, sam}.} uniformly.
Instead, they typically annotate specific event clusters, leaving long timeline segments completely untouched.
The challenge lies in ensuring this localized feedback effectively guides segmentation across the inevitable supervision gaps.
This is a fully supervised setting rather than zero- or few-shot segmentation: models are trained with dense per-timestep state labels, while sparse prompts let experts correct sequence-specific errors and select the desired granularity at inference without retraining.

Prompt-based sliding-window methods such as PromptTSS \cite{prompttss} are \textit{stateless} with respect to user interaction: they apply guidance locally and do not retain it across windows.
Conventional supervised segmenters \cite{prectime, ms_tcn, u_time} are not natively interactive and, when equipped with a stateless prompting interface, likewise cannot persist feedback across windows.
These stateless prompting mechanisms therefore operate under an implicit \textit{Gapless Supervision Assumption}, expecting every sliding window to contain at least one prompt.
In realistic grouped scenarios, vast stretches become \textit{unprompted windows} lacking immediate signals, as shown in Figure~\ref{fig:unprompt_region}.
Because interaction state is local to each window, these methods suffer from \textit{contextual amnesia}: once the window moves past an annotation, guidance is inaccessible \cite{sam_2}.
The trained model still produces predictions in these gaps, but it cannot use the expert's earlier corrections.
Consequently, most of the sequence is segmented without leveraging user guidance, limiting interactive labeling efficiency.

Our analysis reveals this limitation is catastrophic under strictly grouped sparse supervision.
When supervision density drops below 5\% and is concentrated in bursts, the evaluated stateless prompting baselines degrade sharply—dropping from near-perfect accuracy to approximately 60\% in multi-granularity settings.
Without a mechanism to propagate user intent, these models cannot bridge the gap between sparse interaction and global segmentation, limiting their usefulness in grouped-feedback deployments.

\begin{sloppypar}
To bridge these gaps, we propose \textbf{Perseus} (\textbf{Per}sistent \textbf{Se}gmentation with \textbf{U}ser \textbf{S}upervision), a framework transitioning from synchronous stateless prompting to \textit{asynchronous state management}.
Unlike stateless prompting approaches, Perseus decouples supervision from inference via a distinct \textit{Write-Read architecture}.
In the asynchronous write phase, sparse cues are encoded into a persistent memory bank.
During the sequential read phase, the model actively queries this state to service unprompted windows, sustaining intent across unannotated gaps.
By maintaining persistent state, Perseus ensures global consistency regardless of feedback distribution.
\end{sloppypar}

In summary, our contributions are:
\begin{itemize}
\item \textbf{Asynchronous Stateful Framework.}
We propose Perseus, a framework that decouples prompt encoding from inference via an asynchronous \textit{Write-Read} mechanism.
By fusing sparse prompts with local context into a persistent memory bank, Perseus bridges contextual amnesia in stateless prompting.
\item \textbf{Robustness to Grouped Supervision.}
We identify the unprompted window phenomenon as a critical failure mode.
Experiments show that while evaluated stateless baselines degrade significantly when feedback is grouped, Perseus maintains robustness, achieving up to 85\% accuracy improvement in multi-granularity tasks.
\item \textbf{Comprehensive Evaluation.} Evaluating on six wearable and industrial datasets, Perseus demonstrates superior refinement efficiency in iterative settings, achieving an average per-iteration gain of 2.66 percentage points compared to 1.19 for the stateless PromptTSS baseline.
\end{itemize}

\section{Related Work}
\label{sec:related_work}

\subsection{Stateless Prompting in Time Series Segmentation}
Time-series segmentation encompasses several complementary problem formulations.
Change-point detection localizes transitions in a stream \cite{smart_home_1}, while unsupervised latent-state inference and subsequence clustering discover recurring regimes and partition sequences without interactive input \cite{time2state,ticc}.
Interactive machine learning instead places user corrections inside an iterative modeling loop \cite{fails_olsen_iml}.
Perseus connects these lines by retaining sparse user guidance for prompt-based segmentation across inference windows, rather than proposing a new offline discovery or clustering objective.

Within supervised segmentation, neural architectures capture complex spatial and temporal dependencies.
Standard architectures like DeepConvLSTM \cite{deepconvlstm} and U-Time \cite{u_time} model local contexts, while methods like MS-TCN++ \cite{ms_tcn} and PrecTime \cite{prectime} improve boundary precision using multi-stage temporal convolutions.

These architectures can model temporal dependencies within their inputs.
Here, \textit{stateless} refers specifically to interaction history across inference windows, not to general sequence-modeling capacity.
Under this definition, PromptTSS \cite{prompttss}, which introduced sparse prompting for time series segmentation, is stateless: it applies each prompt within its local sliding window and does not retain it after that window is processed.
Under grouped sparse prompting, this stateless interface creates \textit{contextual amnesia}: predictions outside the prompted window cannot access earlier user guidance.

Beyond segmentation, related domains offer complementary solutions.
Hierarchical classification frameworks \cite{hierarchical_1, hierarchical_2, multi_label_1, multi_label_2, multi_label_3} target static data rather than dynamic evolution.
Domain adaptation techniques \cite{adaption_1, adaption_2, adaption_3} address distribution shifts but cannot incorporate specific user feedback during inference.
Similarly, active learning \cite{active_1, active_2, active_3} reduces annotation effort but typically requires retraining, which disrupts real-time interactive sessions.
Perseus targets the narrower problem of persisting sparse inference-time guidance across windows for multi-granularity segmentation.

\subsection{Interactive Mining \& Prompting}
Interactive machine learning aims to integrate human feedback directly into the inference process.
In Natural Language Processing (NLP), Large Language Models (LLMs) adapt to new tasks via textual instructions \cite{chain_of_thought, react}.
Similarly, in Computer Vision, systems like Segment Anything (SAM) \cite{sam, sam_2} allow users to steer segmentation via spatial clicks or boxes.
It is important to note that in the context of time series segmentation, prompting refers to the injection of sparse signal cues (e.g., label clicks or boundary markers) rather than natural language instructions used in LLMs \cite{prompttss}.

For time series, PromptTSS applies prompt signals only within the immediate local region \cite{prompttss}.
Consequently, most of the sequence remains unprompted and segmented without leveraging user guidance.
This locality restricts utility in high-stakes domains like industrial monitoring or healthcare, where the goal is for a small number of expert annotations to propagate consistently across long, complex sequences.
Our work addresses this gap by introducing a stateful memory mechanism that persists encoded prompts, ensuring that interactive guidance extends beyond local neighborhoods to achieve global coherence.

\section{Methodology}
\label{sec:methodology}

\begin{figure*}[t]
    \centering
    \includegraphics[width=2.0\columnwidth]{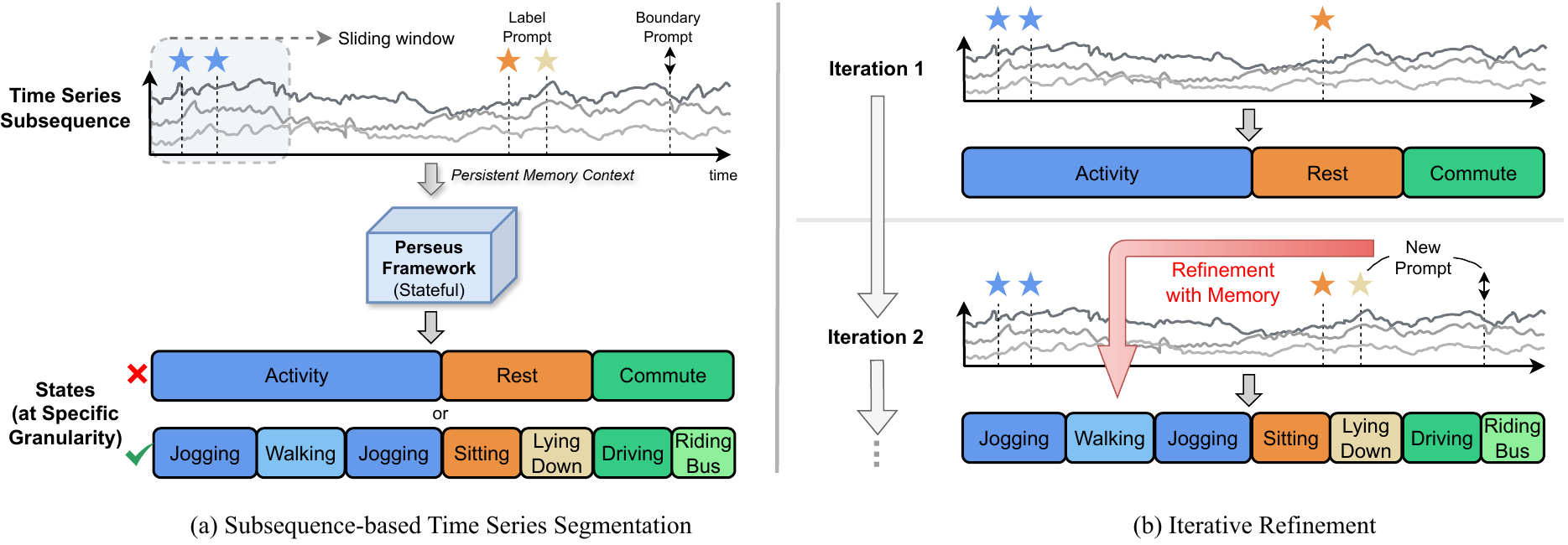}
    \caption{Problem setup and iterative refinement in Perseus. 
    (a) Segmentation is performed over subsequences rather than a single sliding window, with prompts provided as grouped label or boundary cues. 
    (b) In the iterative process, additional prompts are encoded into the persistent memory bank to update the global state.
    This enables the model to refine predictions across the entire timeline, transforming coarse regimes (e.g., Activity, Rest) into fine-grained actions (e.g., Jogging, Sitting), even in regions that received no new local feedback.}
    \label{fig:problem_and_iter}
\end{figure*}

\subsection{Problem Formulation}
\label{sec:problem_formulation}

Figure~\ref{fig:problem_and_iter}(a) illustrates the overall problem formulation.
Let $\mathbf{X} = (x_1, \ldots, x_L) \in \mathbb{R}^{L \times C}$ denote a complete and evenly-sampled multivariate time series of length $L$ with $C$ channels.
The corresponding ground-truth state sequence is $\mathbf{S} = (s_1, \ldots, s_L) \in \mathbb{Z}^L$, where each $s_t$ belongs to one of $K$ discrete states.
To make training and inference efficient, we partition $\mathbf{X}$ into $M$ non-overlapping \textit{subsequences}, each of length $L_s$:  
\[
\mathbf{X} = [\mathbf{x}^{(1)}, \mathbf{x}^{(2)}, \ldots, \mathbf{x}^{(M)}], \quad 
\mathbf{x}^{(m)} \in \mathbb{R}^{L_s \times C}.
\]
Here, $M = \lfloor L / L_s \rfloor$ denotes the total number of subsequences.
Each subsequence $\mathbf{x}^{(m)}$ is associated with its state labels $\mathbf{s}^{(m)} \in \mathbb{Z}^{L_s}$.
Within each subsequence, we further extract $W$ sliding windows of length $T$ and stride $S$:  
\[
\mathbf{x}^{(m)} \mapsto \{ \mathbf{w}^{(m)}_1, \ldots, \mathbf{w}^{(m)}_W \}, \quad
\mathbf{w}^{(m)}_j \in \mathbb{R}^{T \times C}.
\]

\paragraph{\textbf{Prompt Definition}}
In addition to the time series data, we allow sparse prompts to guide segmentation.
A \textit{label prompt} is represented as a vector $p_l \in \{0,1\}^{2K}$, where the first $K$ dimensions encode the correct class in a one-hot manner and the second $K$ dimensions encode incorrect classes using a multi-hot representation.
A \textit{boundary prompt} is represented as a binary value $p_b \in \{0,1\}$, indicating whether a state transition occurs at a given timestamp.
Each prompt corresponds to exactly one timestamp within the subsequence.
We denote the set of all prompts provided for a subsequence $m$ as $\mathcal{P}^{(m)}$.
The goal of time series segmentation is to learn a mapping that, given a subsequence $\mathbf{x}^{(m)}$ together with a set of prompts $\mathcal{P}^{(m)}$, predicts its corresponding state sequence $\mathbf{s}^{(m)}$.

\paragraph{\textbf{Grouped Supervision \& The Gapless Assumption}}
Stateless prompt-based sliding-window models implicitly assume \textit{Gapless Supervision}, where every sliding window is expected to contain at least one prompt ($\forall j, \mathcal{P}^{(m)} \cap \mathbf{w}^{(m)}_j \neq \emptyset$).
However, real-world interaction is grouped and sparse, creating \textit{unprompted windows} where local supervision is absent ($\mathcal{P}^{(m)} \cap \mathbf{w}^{(m)}_j = \emptyset$).
In these gaps, the trained model can still predict from its input, but cannot use guidance provided elsewhere in the subsequence.
Perseus addresses this by learning a stateful mapping $\hat{y} = F(\mathbf{w}^{(m)}_j, \mathbf{M}_{\text{all}})$, where predictions are conditioned on the global history of prompts stored in memory rather than local signals alone.

\subsection{Perseus Architecture}
\label{sec:model_architecture}
\begin{figure*}[t]
    \centering
    \includegraphics[width=2.0\columnwidth]{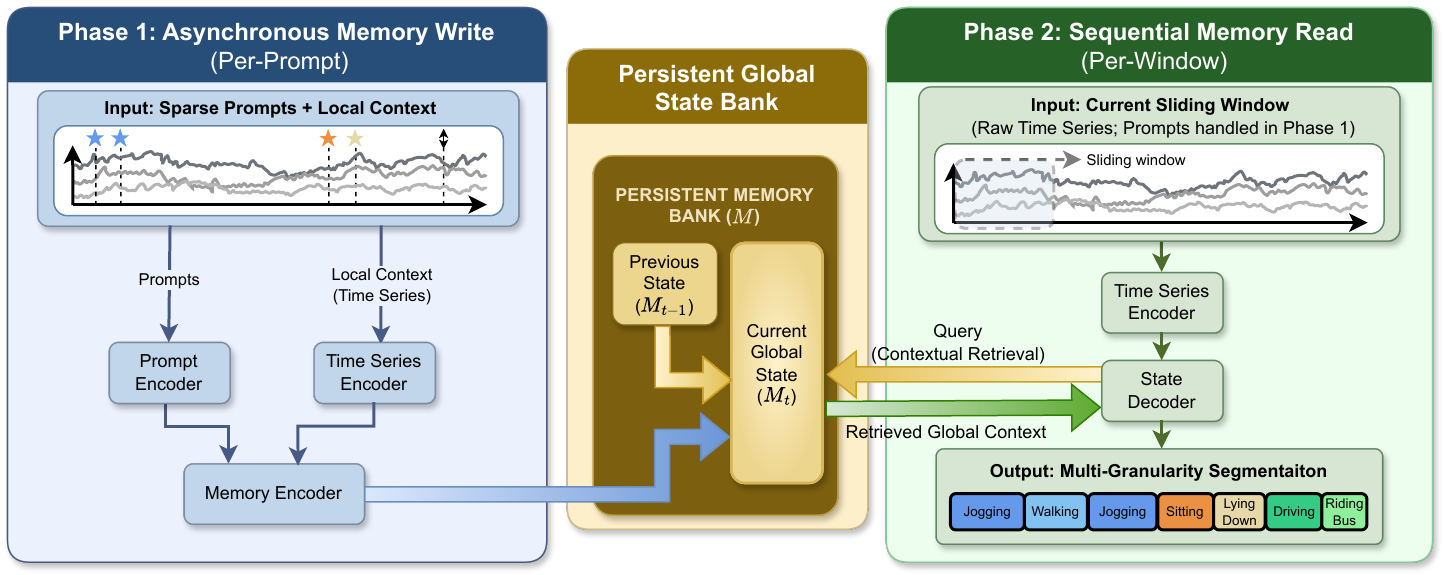}
    \caption{Overview of the \textbf{Perseus} framework featuring asynchronous state management. 
    \textbf{Phase 1 (Memory Write):} Sparse prompts and their local context are asynchronously encoded into a persistent global state bank to update the memory tokens. 
    \textbf{Phase 2 (Memory Read):} During sequential inference, the state decoder actively queries this memory bank to retrieve context for every sliding window, including unprompted ones. 
    This decouples supervision from inference, allowing the model to progressively refine multi-granularity segmentation as the global state evolves.}
    \label{fig:framework}
\end{figure*}

To address the \textit{contextual amnesia} caused by stateless prompting, Perseus introduces \textit{Asynchronous State Management}.
Unlike stateless prompt-based approaches that pair prompts with individual windows, Perseus decouples the lifespan of user feedback from the processing window via a distinct \textit{Write-Read} mechanism.
The framework consists of five main components: the time series encoder, the prompt encoder, the memory encoder, the memory bank, and the state decoder.
Figure~\ref{fig:framework} illustrates the overall architecture, distinguishing between the asynchronous memory write phase (Phase 1) and the sequential memory read phase (Phase 2).

\paragraph{\textbf{Time Series Encoder}}
Given a window $\mathbf{w} \in \mathbb{R}^{T \times C}$, the time series encoder $f_x$ maps it into a sequence of patch-level tokens.
Directly applying a Transformer to long time series windows is computationally expensive.
To address this, we employ \emph{patching}~\cite{patchtst, sam_2}, where adjacent time steps are grouped into patch tokens, reducing sequence length while preserving local temporal patterns.
Formally, the patching operation produces
\begin{equation}
\mathbf{w}_{\text{patched}} = \text{Patching}(\mathbf{w}), 
\end{equation}
where $\mathbf{w}_{\text{patched}} \in \mathbb{R}^{T_p \times (C \cdot P)}$ with patch length $P$ and $T_p$ denoting the number of patches.
The Transformer encoder is then applied:
\begin{equation}
\mathbf{z}_x = f_x(\mathbf{w}_{\text{patched}}), 
\end{equation}
yielding $\mathbf{z}_x \in \mathbb{R}^{T_p \times D}$, where $D$ is the embedding dimension.

\paragraph{\textbf{Prompt Encoder}}
\begin{sloppypar}
The prompt encoder $f_p$ maps each user-provided prompt into a $D$-dimensional embedding.
Each prompt corresponds to exactly one timestamp within a subsequence.
We consider two types of prompts: label prompts and boundary prompts.
\end{sloppypar}

A label prompt is represented as a vector $p_l \in \{0,1\}^{2K}$.
The first $K$ dimensions form a one-hot vector indicating the correct class for the chosen timestamp, while the second $K$ dimensions form a multi-hot vector indicating classes that should not occur at that timestamp.
This dual encoding allows the prompt to convey both positive and negative supervision simultaneously.
A boundary prompt is represented as a binary value $p_b \in \{0,1\}$.
Here, $p_b = 1$ specifies that a state transition should occur at the given timestamp, while $p_b = 0$ specifies that the state should remain unchanged.
Thus, boundary prompts directly guide the model in refining segmentation boundaries.
Both types of prompts are projected into the shared embedding space $\mathbb{R}^D$.
Label prompts are transformed by a linear projection, whereas boundary prompts are mapped through a learned embedding table.
In addition, every prompt embedding is augmented with a type embedding (distinguishing label vs.\ boundary) and an aspect embedding (distinguishing correct vs.\ incorrect).  
The type embedding identifies whether the payload is a label or boundary cue, while the aspect embedding identifies whether that cue is declared correct or incorrect.
After contextual fusion, these distinctions remain encoded in the memory token that the state decoder later queries.
The final prompt embedding is expressed as:
\begin{equation}
\mathbf{z}_p = \begin{cases} 
\text{Linear}(p_l) + \mathbf{e}_{\text{type}} + \mathbf{e}_{\text{aspect}} & \text{if prompt is label} \\ 
\text{Embed}(p_b) + \mathbf{e}_{\text{type}} + \mathbf{e}_{\text{aspect}} & \text{if prompt is boundary} 
\end{cases}
\end{equation}

\paragraph{\textbf{Memory Encoder (Write Mechanism)}}
The memory encoder $f_m$ drives the \emph{asynchronous memory write} phase (Phase 1).
Operating independently of the sliding window inference, it transforms each sparse prompt into a \emph{memory token} that can be persisted in the memory bank.
Its purpose is to fuse the supervision carried by the prompt with local temporal evidence from the subsequence, ensuring that the stored token reflects both the user guidance and the surrounding signal dynamics.
For a prompt anchored at timestamp $t_c$, we extract a local context window of length $T_{ctx}$ centered at $t_c$.
This window is encoded by the time series encoder $f_x$ to produce context tokens $\mathbf{z}_{ctx} \in \mathbb{R}^{T_{ctx} \times D}$.
Given the prompt embedding $\mathbf{z}_p \in \mathbb{R}^{D}$ and the context tokens $\mathbf{z}_{ctx}$, the memory encoder applies cross-attention with the prompt as query and the context as key and value:  
\begin{equation}
\mathbf{m} = f_m(\mathbf{z}_p, \mathbf{z}_{ctx}), 
\end{equation}
where $\mathbf{m} \in \mathbb{R}^{D}$ is the resulting memory token.
This token captures both the user's intent at $t_c$ and the semantic context of the event, preparing it for global retrieval.

\paragraph{\textbf{Memory Bank}}
The memory bank persists memory tokens across iterations, enabling the model to service unprompted windows.
At the beginning of an iteration, it contains previously written tokens $\mathbf{M}_{\text{old}} \in \mathbb{R}^{N_{\text{old}} \times D}$.
New tokens $\mathbf{M}_{\text{cur}} \in \mathbb{R}^{N_p \times D}$, generated by the memory encoder from $N_p$ prompts in the current iteration, are appended to form the updated bank:
\begin{equation}
\mathbf{M}_{\text{all}} = [\mathbf{M}_{\text{old}} ; \mathbf{M}_{\text{cur}}], \quad 
N_{\text{all}} = N_{\text{old}} + N_p,
\end{equation}
where $[\cdot;\cdot]$ denotes concatenation.
By persisting across iterations, the memory bank accumulates knowledge from multiple prompts, supporting long-term refinement of segmentation predictions.

\paragraph{\textbf{State Decoder (Read Mechanism)}}
The state decoder $g_s$ drives the \emph{sequential memory read} phase (Phase 2).
It processes raw sliding windows during inference and actively queries the persistent memory bank to retrieve context, even when the local window contains no prompts.
We employ a \emph{Two-Way Transformer} that maintains a time-series stream $\mathbf{X}$ and a memory stream $\mathbf{M}$.
First, self-attention lets the time-series tokens exchange context within the current window.
Second, the time-series tokens query the memory tokens ($Q=\mathbf{X}$, $K=V=\mathbf{M}$), injecting retrieved supervision into the current-window representation.
Third, self-attention among memory tokens integrates evidence across stored prompts.
Fourth, the memory tokens query the updated time-series tokens ($Q=\mathbf{M}$, $K=V=\mathbf{X}$), aligning the memory stream with the current window.
Fifth, separate feed-forward networks update both streams.
The prediction head reads only the updated time-series stream after the blocks are stacked.
Updates to the memory stream are local to the decoder call and do not overwrite the persistent memory bank.
Formally, given window embeddings $\mathbf{z}_x \in \mathbb{R}^{T_p \times D}$ and memory tokens $\mathbf{M}_{\text{all}} \in \mathbb{R}^{N_{\text{all}} \times D}$ from the memory bank, the decoder produces
\begin{equation}
\hat{\mathbf{y}} = g_s(\mathbf{z}_x, \mathbf{M}_{\text{all}}) \in \mathbb{R}^{T \times K}.
\end{equation}
Here $\hat{\mathbf{y}}$ denotes the predicted per-timestep state probabilities after de-patchification.
Since the encoder operates on patch tokens, de-patchification expands patch-level logits back to the original timestamp resolution by distributing each patch prediction to its covered time steps and averaging in regions of overlap.
This architecture ensures that predictions are informed both by the current input and by the accumulated supervision stored in the memory bank.

\paragraph{\textbf{Complexity \& Scalability Analysis}}
Each prompt contributes one $D$-dimensional memory token, so storage is $O(N_{\text{all}}D)$ and grows linearly with the cumulative prompt count unless the bank is capped.
For each window, cross-attention between $T_p$ patch tokens and $N_{\text{all}}$ memory tokens costs $O(T_pN_{\text{all}}D)$, while self-attention among memory tokens contributes $O(N_{\text{all}}^2D)$.
Treating the embedding dimension as fixed, the additional read cost is therefore $O(T_pN_{\text{all}} + N_{\text{all}}^2)$, not $O(1)$ with respect to the number of stored prompts.
Under a fixed sparse prompt budget, $N_{\text{all}}$ is bounded and this overhead remains modest for a given window length.
A capacity-limited deployment could additionally use an eviction policy, whose accuracy tradeoff we leave for future study.

\subsection{Iterative Training and Evaluation}
\label{sec:training}

Perseus is fully supervised rather than zero- or few-shot: dense ground-truth state sequences supervise the loss at every training timestamp, while sparse prompts simulated from those labels train the interaction pathway.
At inference, model weights remain fixed and users provide only sparse label or boundary prompts to correct sequence-specific errors and express the desired granularity without retraining.
The iterative procedure progressively incorporates these prompts and accumulates them in memory, modeling how segmentation improves as feedback is added (Figure~\ref{fig:problem_and_iter}(b)).
At each iteration $r$, the model performs two key procedures:

\paragraph{\textbf{Step 1: Prompt Sampling \& State Update}}
At the start of iteration $r$, we simulate interactive feedback by sampling $N_p$ new prompts from the ground-truth labels for each subsequence.
Unlike the window-based inference, this sampling occurs at the \emph{subsequence level}.
For a subsequence $m$, the sampled prompts are encoded ($f_p$) and fused with local context ($f_m$) to generate new memory tokens:
\begin{equation}
\mathbf{M}_{\text{cur}}^{(m)} = f_m(f_p(p^{(m)}), \mathbf{z}_{ctx}^{(m)}).
\end{equation}
These new tokens $\mathbf{M}_{\text{cur}}^{(m)}$ are immediately appended to the memory bank, updating the global state available for the subsequent inference step.

\paragraph{\textbf{Step 2: Optimization via Memory Read}}
Once the state is updated, we perform inference on every sliding window $\mathbf{w}^{(m)}_j$ within the subsequence.
All prompts collected in iteration $r$ are written before any window is read, so our experiments model offline, non-causal refinement: a prompt later on the timeline can guide an earlier window within the same subsequence.
The state decoder $g_s$ utilizes the updated memory bank $\mathbf{M}_{\text{all}}^{(m)}$ to generate predictions for unprompted windows:
\begin{equation}
\hat{\mathbf{y}}^{(m)}_{j} = g_s\!\big(\mathbf{z}_{x,j}^{(m)}, \mathbf{M}_{\text{all}}^{(m)}\big).
\end{equation}
The model is optimized using cross-entropy loss averaged over all windows and timesteps in the iteration:
\begin{equation}
\mathcal{L}^{(r)} = \frac{1}{W \cdot T} 
\sum_{j=1}^W \sum_{t=1}^T 
\mathcal{L}_{CE}\!\big(s^{(m)}_{j,t}, \,\hat{y}^{(m)}_{j,t}\big),
\end{equation}
where $s^{(m)}_{j,t}$ is the ground-truth state. 
After the backpropagation step, the new tokens $\mathbf{M}_{\text{cur}}^{(m)}$ are detached from the computation graph and stored in the memory bank for the next iteration.

\section{Experiments}
\label{sec:experiments}

\renewcommand{\arraystretch}{1.20}  % Row spacing (default: 1)
\setlength{\tabcolsep}{2pt}       % Col spacing (default: 6pt)

\begin{table}[t]
% \tiny
\centering
\caption{Statistical overview of datasets used in Perseus experiments.}
\label{tab:dataset}
\resizebox{\linewidth}{!}{
\begin{tabular}{c|cccccc}
\hline
Datasets & \# Features & \# Timesteps & \multicolumn{1}{c}{\# Time Series} & \# States & State Duration & Avg State Duration \\ \hline
Pump V35 & 9 & 27,770 $\sim$ 40,810 & \multicolumn{1}{c}{40} & 41 $\sim$ 43 & 1 $\sim$ 3,290 & 543 \\
Pump V36 & 9 & 26,185 $\sim$ 38,027 & \multicolumn{1}{c}{40} & 41 $\sim$ 43 & 1 $\sim$ 1,960 & 517 \\
Pump V38 & 9 & 20,365 $\sim$ 30,300 & \multicolumn{1}{c}{40} & 42 $\sim$ 43 & 1 $\sim$ 1,820 & 407 \\
% MoCap & 4 & 4,579 $\sim$ 10,617 & \multicolumn{1}{c}{9} & 3 $\sim$ 9 & 354 $\sim$ 2,003 & 886 \\
% ActRecTut & 23 & 31,392 $\sim$ 32,577 & \multicolumn{1}{c}{2} & 6 & 22 $\sim$ 5,100 & 743 \\
USC-HAD & 6 & 25,356 $\sim$ 56,251 & \multicolumn{1}{c}{70} & 12 & 600 $\sim$ 13,500 & 3,347 \\
PAMAP2 & 9 & 8,477 $\sim$ 447,000 & \multicolumn{1}{c}{9} & 2 $\sim$ 13 & 1 $\sim$ 42,995 & 14,434 \\
IndustryMG (Fine-Grained) & 3 & 13,629 $\sim$ 45,010 & \multicolumn{1}{c}{17} & 4 & 10 $\sim$ 2,236 & 583 \\ \hline
Pump V35 (2x Coarser) & \multicolumn{1}{r}{----------} & Same as PumpV35 & \multicolumn{1}{l}{----------} & 22 & 1 $\sim$ 3,570 & 771 \\
Pump V36 (2x Coarser) & \multicolumn{1}{r}{----------} & Same as PumpV36 & \multicolumn{1}{l}{----------} & 21 $\sim$ 22 & 1 $\sim$ 2,085 & 719 \\
Pump V38 (2x Coarser) & \multicolumn{1}{r}{----------} & Same as PumpV38 & \multicolumn{1}{l}{----------} & 22 & 1 $\sim$ 3,305 & 625 \\
USC-HAD (2x Coarser) & \multicolumn{1}{r}{----------} & Same as USC-HAD & \multicolumn{1}{l}{----------} & 6 & 1,700 $\sim$ 19,100 & 6,694 \\
PAMAP2 (2x Coarser) & \multicolumn{1}{r}{----------} & Same as PAMAP2 & \multicolumn{1}{l}{----------} & 6 & 1 $\sim$ 50,198 & 16,997 \\
IndustryMG (Coarse-Grained) & \multicolumn{1}{r}{----------} & \begin{tabular}[c]{@{}c@{}}Same as IndustryMG\\ (Fine-Grained)\end{tabular} & \multicolumn{1}{l}{----------} & 2 & 18 $\sim$ 3,415 & 883 \\ \hline
Pump V35 (4x Coarser) & \multicolumn{1}{r}{----------} & Same as PumpV35 & \multicolumn{1}{l}{----------} & 11 & 1 $\sim$ 3,855 & 1,171 \\
Pump V36 (4x Coarser) & \multicolumn{1}{r}{----------} & Same as PumpV36 & \multicolumn{1}{l}{----------} & 11 & 1 $\sim$ 4,874 & 1,019 \\
Pump V38 (4x Coarser) & \multicolumn{1}{r}{----------} & Same as PumpV38 & \multicolumn{1}{l}{----------} & 11 & 1 $\sim$ 4,154 & 933 \\ \hline
\end{tabular}
}
\end{table}

\paragraph{\textbf{Datasets}} 
We evaluate Perseus on six datasets, with their statistics summarized in Table~\ref{tab:dataset}.
The first group consists of publicly available benchmarks including \textbf{USC-HAD} \cite{usc_had}, \textbf{PAMAP2} \cite{pamap2}, and the \textbf{Pump} datasets (V35, V36, V38) \cite{prectime}.
USC-HAD records human daily activities using wearable accelerometer and gyroscope sensors, while PAMAP2 contains multimodal sensor data collected from subjects performing diverse physical activities.
The Pump datasets are collected from hydraulic pump end-of-line (EoL) testing, each annotated with more than 40 operational states.
These datasets were originally annotated with fine-grained states only, and we create multiple levels of granularity by systematically merging \emph{neighboring states in temporal order}.
For USC-HAD (12 states) and PAMAP2 (24 states), the relatively small number of classes allows only a 2$\times$ coarser version.
For Pump V35, V36, and V38, we construct both 2$\times$ and 4$\times$ coarser levels to capture broader operating phases.
In contrast, \textbf{IndustryMG} is a proprietary industrial dataset with \emph{naturally annotated} multi-granularity states, capturing both high-level production phases and fine-grained machine operations.
Unlike the synthetic coarsening applied to the open datasets, IndustryMG directly provides real-world hierarchical annotations, making it an important benchmark for validating adaptive segmentation in practice.
IndustryMG cannot be redistributed because the measurements were supplied under third-party confidentiality restrictions.
No proprietary records are included in the public release.
Code, preprocessing instructions, and the open-data pipeline are available at \url{https://github.com/blacksnail789521/Perseus}.

\renewcommand{\arraystretch}{1.15}  % Row spacing (default: 1)
\setlength{\tabcolsep}{6pt}       % Col spacing (default: 6pt)

\begin{table*}
\centering
\caption{Segmentation performance under the single iteration inference setting.
A budget of $\lfloor0.05L_s\rfloor$ prompt cues is provided once per subsequence.
Best results are in {\color[HTML]{FF0000} \textbf{bold}}, and second-best are {\color[HTML]{0000FF} {\ul underlined}}.}
\label{tab:single_iter}
\resizebox{\textwidth}{!}{%
\begin{tabular}{cc|cc|cc|cc|cc|cc|cc}
\hline
\multicolumn{2}{c|}{Dataset} & \multicolumn{2}{c|}{USC-HAD} & \multicolumn{2}{c|}{PAMAP2} & \multicolumn{2}{c|}{IndustryMG} & \multicolumn{2}{c|}{Pump V35} & \multicolumn{2}{c|}{Pump V36} & \multicolumn{2}{c}{Pump V38} \\ \hline
\multicolumn{2}{c|}{Granularity} & Single & Multiple & Single & Multiple & Single & Multiple & Single & Multiple & Single & Multiple & Single & Multiple \\ \hline
\multicolumn{1}{c|}{} & ACC & {\color[HTML]{FE0000} \textbf{92.70}} & {\color[HTML]{FE0000} \textbf{95.56}} & {\color[HTML]{FE0000} \textbf{94.94}} & {\color[HTML]{FE0000} \textbf{96.96}} & 96.32 & {\color[HTML]{FE0000} \textbf{89.68}} & {\color[HTML]{FE0000} \textbf{88.96}} & {\color[HTML]{FE0000} \textbf{92.34}} & {\color[HTML]{FE0000} \textbf{91.61}} & {\color[HTML]{FE0000} \textbf{91.83}} & {\color[HTML]{FE0000} \textbf{85.55}} & {\color[HTML]{FE0000} \textbf{88.95}} \\
\multicolumn{1}{c|}{} & MF1 & {\color[HTML]{FE0000} \textbf{91.18}} & {\color[HTML]{FE0000} \textbf{93.92}} & {\color[HTML]{FE0000} \textbf{93.51}} & {\color[HTML]{FE0000} \textbf{96.16}} & 79.50 & {\color[HTML]{FE0000} \textbf{71.58}} & {\color[HTML]{FE0000} \textbf{82.35}} & {\color[HTML]{FE0000} \textbf{88.33}} & {\color[HTML]{FE0000} \textbf{86.38}} & {\color[HTML]{FE0000} \textbf{85.94}} & {\color[HTML]{FE0000} \textbf{79.61}} & {\color[HTML]{FE0000} \textbf{83.08}} \\
\multicolumn{1}{c|}{\multirow{-3}{*}{Perseus (Ours)}} & ARI & {\color[HTML]{FE0000} \textbf{87.65}} & {\color[HTML]{FE0000} \textbf{91.75}} & {\color[HTML]{FE0000} \textbf{89.32}} & {\color[HTML]{FE0000} \textbf{92.35}} & 92.58 & {\color[HTML]{FE0000} \textbf{79.23}} & {\color[HTML]{FE0000} \textbf{86.03}} & {\color[HTML]{FE0000} \textbf{86.99}} & {\color[HTML]{FE0000} \textbf{89.69}} & {\color[HTML]{FE0000} \textbf{87.14}} & {\color[HTML]{FE0000} \textbf{78.53}} & {\color[HTML]{FE0000} \textbf{80.97}} \\ \hline
\multicolumn{1}{c|}{} & ACC & {\color[HTML]{0000FF} {\ul 80.53}} & {\color[HTML]{0000FF} {\ul 60.91}} & 51.90 & {\color[HTML]{0000FF} {\ul 53.54}} & 94.79 & 71.60 & {\color[HTML]{0000FF} {\ul 70.93}} & {\color[HTML]{0000FF} {\ul 41.47}} & {\color[HTML]{0000FF} {\ul 77.01}} & {\color[HTML]{0000FF} {\ul 41.69}} & {\color[HTML]{0000FF} {\ul 72.33}} & {\color[HTML]{0000FF} {\ul 41.98}} \\
\multicolumn{1}{c|}{} & MF1 & {\color[HTML]{0000FF} {\ul 77.30}} & {\color[HTML]{0000FF} {\ul 52.01}} & 47.22 & {\color[HTML]{0000FF} {\ul 41.09}} & 72.05 & 40.27 & 54.20 & {\color[HTML]{0000FF} {\ul 32.26}} & {\color[HTML]{0000FF} {\ul 63.61}} & {\color[HTML]{0000FF} {\ul 32.53}} & 55.45 & {\color[HTML]{0000FF} {\ul 34.14}} \\
\multicolumn{1}{c|}{\multirow{-3}{*}{PromptTSS}} & ARI & {\color[HTML]{0000FF} {\ul 68.23}} & {\color[HTML]{0000FF} {\ul 39.60}} & 31.09 & 11.55 & 89.24 & 55.24 & {\color[HTML]{0000FF} {\ul 62.06}} & 20.74 & {\color[HTML]{0000FF} {\ul 68.55}} & {\color[HTML]{0000FF} {\ul 20.96}} & {\color[HTML]{0000FF} {\ul 64.30}} & {\color[HTML]{0000FF} {\ul 21.06}} \\ \hline
\multicolumn{1}{c|}{} & ACC & 72.48 & 51.21 & 44.19 & 44.00 & 96.46 & 77.10 & 62.05 & 31.38 & 60.46 & 31.71 & 63.93 & 33.02 \\
\multicolumn{1}{c|}{} & MF1 & 68.15 & 46.03 & 31.23 & 27.96 & 72.49 & 66.12 & {\color[HTML]{0000FF} {\ul 56.25}} & 29.66 & 57.78 & 29.23 & {\color[HTML]{0000FF} {\ul 57.13}} & 30.44 \\
\multicolumn{1}{c|}{\multirow{-3}{*}{PrecTime}} & ARI & 58.35 & 30.80 & 13.23 & 13.55 & 92.57 & 67.37 & 59.81 & 17.65 & 50.26 & 16.36 & 55.96 & 17.69 \\ \hline
\multicolumn{1}{c|}{} & ACC & 31.41 & 26.29 & 43.00 & 43.82 & 96.75 & 76.97 & 54.18 & 28.39 & 41.71 & 23.08 & 37.99 & 21.98 \\
\multicolumn{1}{c|}{} & MF1 & 22.72 & 16.76 & 18.28 & 19.50 & 70.25 & 63.36 & 20.11 & 14.42 & 15.18 & 14.62 & 10.65 & 10.64 \\
\multicolumn{1}{c|}{\multirow{-3}{*}{MS-TCN++}} & ARI & 19.06 & 14.28 & 9.10 & 19.83 & {\color[HTML]{FE0000} \textbf{95.49}} & 68.46 & 59.19 & 18.75 & 38.94 & 13.28 & 28.91 & 10.55 \\ \hline
\multicolumn{1}{c|}{} & ACC & 48.24 & 32.11 & 45.32 & 41.07 & {\color[HTML]{FE0000} \textbf{97.69}} & 77.53 & 43.01 & 16.85 & 25.03 & 12.58 & 49.06 & 27.23 \\
\multicolumn{1}{c|}{} & MF1 & 47.21 & 25.99 & 31.38 & 22.43 & {\color[HTML]{0000FF} {\ul 90.07}} & 63.99 & 23.03 & 11.99 & 12.17 & 7.06 & 23.14 & 18.01 \\
\multicolumn{1}{c|}{\multirow{-3}{*}{U-Time}} & ARI & 26.62 & 15.73 & 26.27 & 21.78 & {\color[HTML]{FE0000} \textbf{95.49}} & 68.97 & 27.55 & 6.71 & 15.56 & 3.75 & 42.30 & 15.27 \\ \hline
\multicolumn{1}{c|}{} & ACC & 32.81 & 30.65 & {\color[HTML]{0000FF} {\ul 58.15}} & 48.65 & {\color[HTML]{FE0000} \textbf{97.69}} & {\color[HTML]{0000FF} {\ul 77.84}} & 63.78 & 35.06 & 61.32 & 34.31 & 56.01 & 30.72 \\
\multicolumn{1}{c|}{} & MF1 & 34.23 & 27.39 & {\color[HTML]{0000FF} {\ul 47.86}} & 29.54 & {\color[HTML]{FE0000} \textbf{90.34}} & {\color[HTML]{0000FF} {\ul 66.17}} & 48.13 & 28.78 & 46.52 & 31.02 & 31.94 & 24.39 \\
\multicolumn{1}{c|}{\multirow{-3}{*}{DeepConvLSTM}} & ARI & 18.77 & 12.73 & {\color[HTML]{0000FF} {\ul 32.35}} & {\color[HTML]{0000FF} {\ul 31.19}} & 95.43 & {\color[HTML]{0000FF} {\ul 69.16}} & 60.53 & {\color[HTML]{0000FF} {\ul 21.90}} & 50.75 & 18.44 & 50.02 & 17.70 \\ \hline
\multicolumn{1}{c|}{} & ACC & 29.28 & 30.05 & 36.32 & 36.50 & 67.25 & 55.99 & 27.44 & 16.72 & 40.96 & 21.73 & 32.17 & 18.24 \\
\multicolumn{1}{c|}{} & MF1 & 16.46 & 15.21 & 8.12 & 7.89 & 29.60 & 32.31 & 11.80 & 9.39 & 15.72 & 12.80 & 11.09 & 10.19 \\
\multicolumn{1}{c|}{\multirow{-3}{*}{iTransformer-TSS}} & ARI & 14.04 & 10.51 & 9.58 & 11.49 & 30.26 & 18.84 & 19.32 & 6.76 & 31.32 & 10.01 & 19.73 & 6.41 \\ \hline
\multicolumn{1}{c|}{} & ACC & 60.74 & 45.77 & 38.86 & 38.21 & 50.32 & 31.52 & 61.61 & 33.09 & 65.73 & 34.03 & 63.05 & 31.97 \\
\multicolumn{1}{c|}{} & MF1 & 51.35 & 38.33 & 23.95 & 11.92 & 27.11 & 19.83 & 42.70 & 26.01 & 43.39 & 26.74 & 40.12 & 23.56 \\
\multicolumn{1}{c|}{\multirow{-3}{*}{PatchTST-TSS}} & ARI & 45.50 & 27.74 & 13.31 & 4.81 & 16.74 & 6.29 & 55.74 & 17.95 & 55.75 & 17.54 & 55.06 & 16.43 \\ \hline
\end{tabular}%
}
\end{table*}

\paragraph{\textbf{Metrics}} 
To comprehensively evaluate segmentation performance, we adopt three commonly used metrics: 
Accuracy (ACC), Macro F1-score (MF1), and the Adjusted Rand Index (ARI).
Accuracy measures the overall proportion of correctly predicted states across the sequence.
Macro F1-score balances performance across classes by averaging the F1-score of each state, which is particularly important in time series with imbalanced state distributions.
Adjusted Rand Index evaluates clustering consistency by comparing predicted and ground-truth state assignments while adjusting for chance, providing a robust measure of segmentation quality.
Metrics are computed over the windowed timestamp instances produced by the evaluation loader: if a timestamp occurs in several overlapping windows, each occurrence is counted once per window rather than aggregated into one subsequence-level prediction.
This overlap-weighted protocol is applied identically to every method and matches the training unit, but it is not a one-vote-per-unique-timestamp evaluation.
We state this distinction to avoid implying otherwise.
Boundary errors are therefore reflected through timestamp labels rather than a relaxed boundary tolerance.

\begin{figure}[t]
    \centering
    \includegraphics[width=1.0\columnwidth]{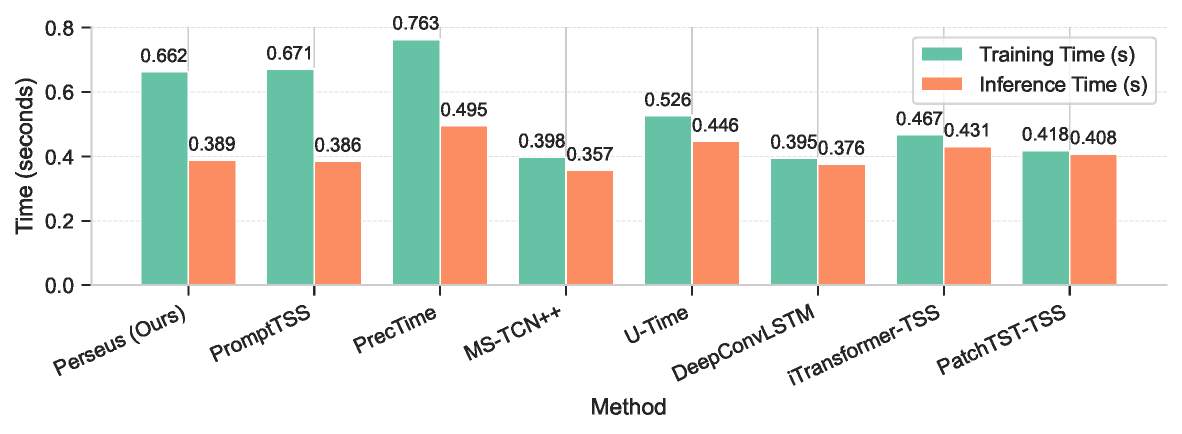}
    \caption{Comparison of training and inference times per batch on the PAMAP2 dataset.}
    \label{fig:running_time}
\end{figure}

\paragraph{\textbf{Baselines}} 
We compare Perseus against a broad range of state-of-the-art segmentation models.
Throughout this evaluation, \textit{stateless} refers specifically to interaction history: none of the baselines retains prompted feedback across inference windows.
\textbf{PromptTSS} \cite{prompttss} represents the state-of-the-art in prompting-based segmentation.
It incorporates label and boundary cues but operates under a local sliding-window assumption.
While highly effective under uniform supervision, it lacks subsequence-level persistence, making it vulnerable to grouped supervision gaps. 
\textbf{PrecTime} \cite{prectime} is a sequence-to-sequence architecture combining sliding windows with dense labeling, designed for precise industrial segmentation but limited to local context. 
\textbf{MS-TCN++} \cite{ms_tcn} applies multi-stage temporal convolutional layers with dilations to refine predictions and mitigate over-segmentation. 
\textbf{U-Time} \cite{u_time} adapts the U-Net architecture for 1D time series, capturing both local and global temporal context within a fixed window. 
\textbf{DeepConvLSTM} \cite{deepconvlstm} combines convolutional feature extraction with recurrent modeling for wearable activity recognition. 
Finally, we include \textit{forecasting adapters} (\textbf{iTransformer-TSS} \cite{itransformer}, \textbf{PatchTST-TSS} \cite{patchtst}), which are segmentation variants of strong forecasting models.
While they capture long-range dependencies, they remain stateless with respect to user interaction history across inference passes.

Among all baselines, only Perseus and PromptTSS support prompting natively.
For the remaining baselines, we use a post-hoc prompting wrapper: separate models are trained for each granularity level, each model predicts the full subsequence, and the candidate with the lowest mismatch on the available prompted timestamps is selected.
No ground-truth labels outside the prompted positions are used during inference, and all methods receive the same prompt budget and locations.
We also include a simple \textbf{LabelProp} sanity baseline: under the same hotspot prompt protocol, it copies the nearest declared-correct label prompt through the subsequence.
Boundary prompts and incorrect-label prompts are counted in the budget but are not converted into oracle labels.

\begin{figure*}[t]
    \centering
    \includegraphics[width=2.0\columnwidth]{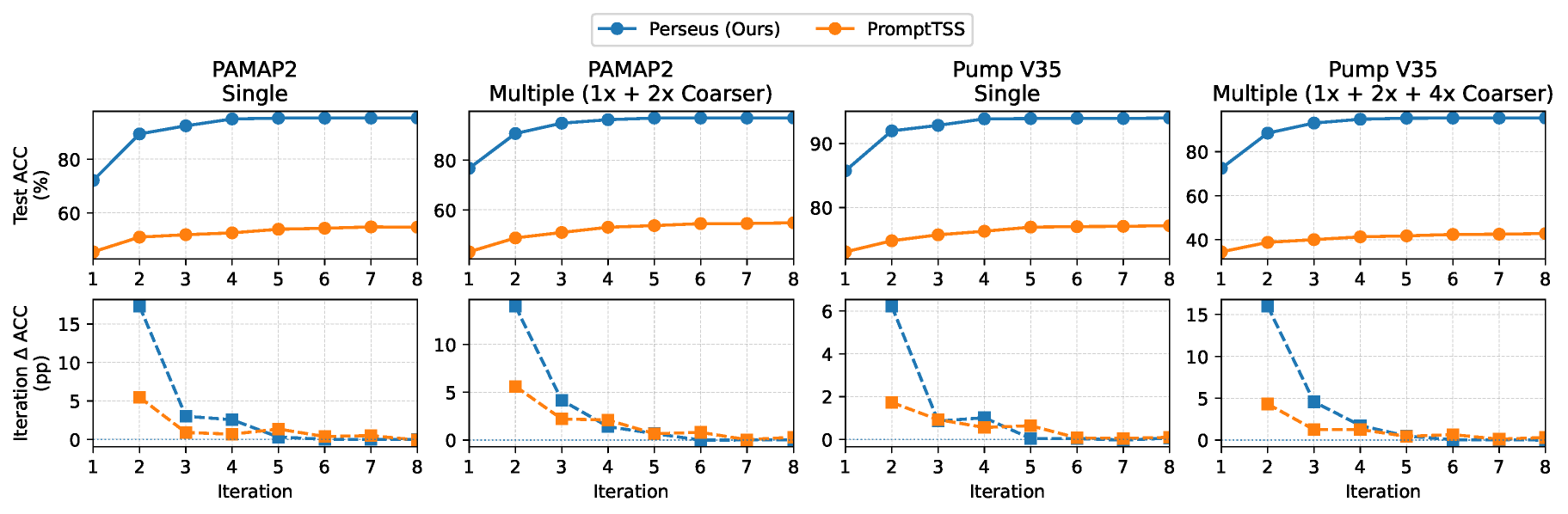}
    \caption{Segmentation performance under the multiple-iteration inference setting, evaluated on two datasets (PAMAP2 and Pump V35), each with both single and multiple granularities of states.
\textbf{Top row:} Test accuracy (ACC, \%). 
\textbf{Bottom row:} Iteration $\Delta$ ACC (percentage points, pp).}
    \label{fig:iterative_infer}
\end{figure*}

\paragraph{\textbf{Experimental Setup: Grouped Sparse Supervision}} 
A critical distinction of our evaluation is the supervision setting.
Unlike prior works that evaluate under dense or uniform prompting assumptions, we use a \textit{grouped sparse budget} and distinguish three protocols below.
For the single-step experiment, the number of prompt cues is $\lfloor0.05L_s\rfloor$, i.e., a budget equal to 5\% of the subsequence length rather than necessarily 5\% distinct prompted timestamps.
Prompts are not uniformly distributed: given $W=8$ windows per subsequence, their timestamps are constrained to two \textit{bursty} windows, leaving the remaining six windows entirely \emph{unprompted}.
This setup mimics realistic user behavior where annotators mark clusters of events and leave large gaps.
Under this setting, we test whether models suffer from contextual amnesia in unprompted windows or can successfully retrieve supervision provided elsewhere in the subsequence.
Prompt simulation draws label and boundary cues with equal probability and correct versus incorrect aspects with equal probability.
For a declared-incorrect label cue, the simulator uniformly samples one of the $K-1$ classes different from the ground-truth class.
For an incorrect boundary cue, it flips the ground-truth boundary bit.

\paragraph{\textbf{Implementation Details}} 
We split each dataset chronologically into 70\% training, 15\% validation, and 15\% testing.
For subsequence construction, we use non-overlapping segments of length $L_s$, each further divided into $W=8$ sliding windows of length $T$ with stride $S$.
We set $T=256$ (or $T=512$ for USC-HAD) with stride $S=T/4$.
Perseus employs iterative prompting at the subsequence level.
During training, each of $N_r=8$ iterations adds $N_p=4$ prompts, for 32 cumulative prompts.
This training budget is distinct from the single-step 5\% test budget.
All models are trained with the AdamW optimizer (learning rate $1\!\times\!10^{-4}$).

\subsection{Single Iteration Inference}
In this setting, user feedback is provided once using the grouped budget of $\lfloor0.05L_s\rfloor$ prompt cues.
This setup serves as a stress test for \textit{state persistence}: can the model retain user intent across the six of eight windows (75\%) that receive no prompt?
The results are summarized in Table~\ref{tab:single_iter}.

\paragraph{\textbf{Overcoming Contextual Amnesia}}
Under this grouped supervision, Perseus achieves an average accuracy improvement of 54\% (23\% on single-granularity and 85\% on multi-granularity datasets) compared to the best-performing neural baseline.
Crucially, the performance of the stateless PromptTSS baseline drops significantly compared to its theoretical ceiling.
This degradation highlights \textit{contextual amnesia} in local prompt-based models: in unprompted windows (inter-burst gaps), PromptTSS cannot access supervision provided in neighboring windows.
In contrast, Perseus utilizes its memory bank to bridge these gaps.
By encoding the "burst" of prompts into persistent memory tokens, it allows unprompted windows to attend to the supervised history.
The massive gain in multi-granularity tasks (117\% improvement over neural baselines excluding PromptTSS) further validates that hierarchical state definitions require global context that only a stateful memory can provide.

\begin{table}[t]
\centering
\caption{Matched label-propagation sanity check on industrial datasets.
\(\Delta\) ACC (pp) is Perseus (ours) ACC minus LabelProp ACC.
Higher is better for all metrics.}
\label{tab:label_prop_sanity}
\scriptsize
\setlength{\tabcolsep}{3.5pt}
\renewcommand{\arraystretch}{1.15}
\begin{tabular}{lc|ccc|ccc|c}
\toprule
\multirow{2}{*}{Dataset} & \multirow{2}{*}{Gran.} & \multicolumn{3}{c|}{LabelProp} & \multicolumn{3}{c|}{Perseus (ours)} & \multirow{2}{*}{\(\Delta\) ACC (pp)} \\
 & & ACC & MF1 & ARI & ACC & MF1 & ARI & \\
\midrule
Pump V36 & Single & 65.79 & 45.75 & 59.83 & 91.61 & 86.38 & 89.69 & {\color[HTML]{FE0000}\textbf{+25.83}} \\
Pump V36 & Multi. & 72.76 & 58.61 & 58.64 & 91.83 & 85.94 & 87.14 & {\color[HTML]{FE0000}\textbf{+19.07}} \\
Pump V38 & Single & 61.29 & 36.57 & 49.73 & 85.55 & 79.61 & 78.53 & {\color[HTML]{FE0000}\textbf{+24.26}} \\
Pump V38 & Multi. & 69.63 & 54.32 & 53.23 & 88.95 & 83.08 & 80.97 & {\color[HTML]{FE0000}\textbf{+19.32}} \\
\bottomrule
\end{tabular}
\end{table}

\paragraph{\textbf{Beyond Label Propagation}}
Table~\ref{tab:label_prop_sanity} evaluates whether a simple persistent-label rule is sufficient.
LabelProp is intentionally strong for a state-machine baseline: it receives the same grouped prompt budget and copies the nearest usable correct-label prompt across unprompted regions.
On Pump V36 and V38, Perseus improves ACC by 19.07--25.83 percentage points and also yields large MF1/ARI gains, showing that industrial multi-state transitions cannot be recovered by copying a single local label.
This result should be interpreted with dataset structure in mind: on activity datasets such as PAMAP2 and USC-HAD, many test subsequences contain one long state, making label propagation competitive.
For example, the median state run spans the full subsequence on PAMAP2 and USC-HAD, with 90.9\% and 67.1\% single-state subsequences in their single-granularity test splits.
The Pump datasets contain shorter state runs and more transitions, making them a stricter test of whether memory can bridge unprompted state changes.

\paragraph{\textbf{Efficiency}}
We further analyze training and inference time for all methods, as shown in Figure~\ref{fig:running_time}.
Despite the addition of the memory bank, Perseus maintains a runtime profile highly similar to PromptTSS.
This indicates that the cost of statefulness is modest under the evaluated sparse prompt budget.
The Write/Read operations use compact token representations rather than high-dimensional feature maps, allowing Perseus to address contextual amnesia while retaining a runtime profile close to PromptTSS in this setting.

\subsection{Iterative Inference}
We next evaluate the dynamic capability of Perseus in an interactive setting using PAMAP2 and Pump V35.
We limit the comparison to Perseus and PromptTSS, the only evaluated methods with native iterative prompt input.
This is a separate fixed-count protocol: each of $N_r=8$ iterations adds four new prompts, yielding 32 cumulative prompts rather than stopping at the single-step 5\% budget.
The results are illustrated in Figure~\ref{fig:iterative_infer}.

\paragraph{\textbf{State Accumulation vs. Stateless Re-computation}}
While both models improve with more feedback, the mechanism of improvement differs fundamentally.
PromptTSS treats each iteration as a largely independent inference pass.
It benefits from new prompts only because the local sliding windows happen to capture them.
Perseus, however, treats iterations as \textit{state accumulation}.
Results show that Perseus achieves an average per-iteration accuracy gain ($\Delta$ ACC) of 2.66 percentage points, compared to just 1.19 for PromptTSS.
The disparity is most critical in the early phase (Iterations 1--4), where Perseus achieves a massive 6.07 pp gain compared to 2.25 pp for the baseline.
This confirms that Perseus is a more \textit{data-efficient learner}: by retrieving previously stored tokens from the memory bank, it compounds the value of every new prompt, whereas the stateless baseline struggles to propagate sparse signals beyond their immediate context.

\begin{figure}[t]
    \centering
    \includegraphics[width=1.0\columnwidth]{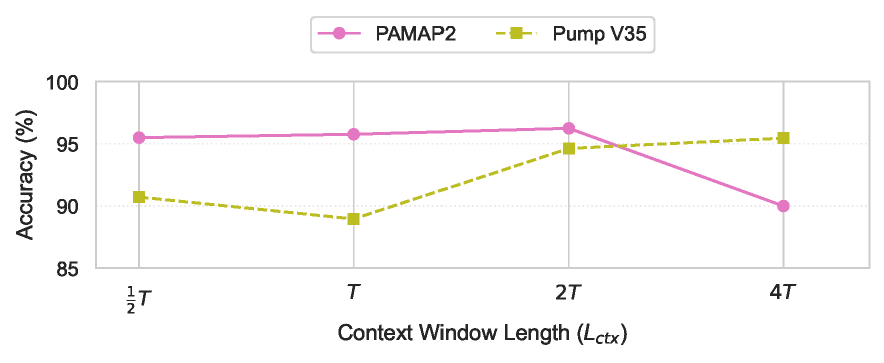}
    \caption{Ablation study on the impact of context window length \(T_{ctx}\) on the segmentation accuracy of Perseus (PAMAP2 and PumpV35).}
    \label{fig:context_window_length}
\end{figure}

\subsection{Ablation Study} 

\paragraph{\textbf{Effect of Context Window Length}} 
We first examine the effect of the context window length $T_{\text{ctx}}$ used when encoding prompts into memory tokens. 
Experiments are conducted on PAMAP2 and Pump V35, where we vary $T_{\text{ctx}}$ across four values: $\tfrac{1}{2}T$, $T$, $2T$, and $4T$. 
The results are shown in Figure~\ref{fig:context_window_length}. 

We observe that increasing $T_{\text{ctx}}$ generally improves segmentation performance, confirming that the model benefits from leveraging more surrounding time series context when generating memory tokens. 
However, the improvement is not permanent. 
On PAMAP2, performance begins to decrease when $T_{\text{ctx}}$ reaches $4T$, suggesting that excessively long context windows may introduce noise or dilute the local temporal cues that are most relevant for prompt alignment. 
This indicates that while richer context is useful for constructing memory, it should remain within a reasonable range to balance local precision and global coverage.

\begin{figure}[t]
    \centering
    \includegraphics[width=1.0\columnwidth]{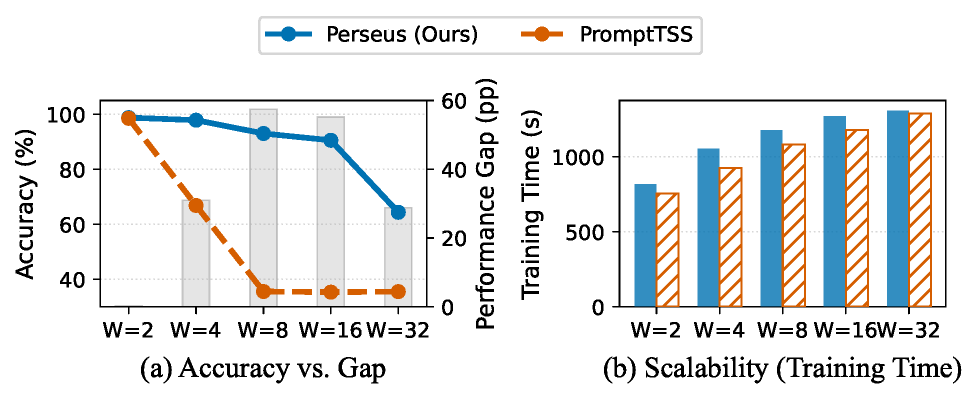}
    \caption{Impact of subsequence length $L_s$ on PAMAP2 (fixed budget). (a) Accuracy and the widening performance delta ($\Delta$) highlight Perseus's superior robustness to signal dilution. (b) Training time shows favorable empirical scaling over the evaluated range.}
    \label{fig:subsequence_length}
\end{figure}

\paragraph{\textbf{Effect of Subsequence Length \& Scalability}}
We next evaluate the model's robustness and scalability by varying the subsequence length $L_s$ (containing $W \in \{2, 4, 8, 16, 32\}$ sliding windows).
To simulate realistic interaction constraints, we adopt a \emph{Fixed User Budget} protocol: we keep the total supervision budget fixed at 32 prompts (accumulated via 8 iterations with $N_p=4$), concentrated within exactly two sliding windows regardless of the total sequence length.
This creates a challenging scenario where supervision density dilutes as $L_s$ increases: coverage drops from 100\% (2 out of 2 windows) at $W=2$ to merely 6.25\% (2 out of 32 windows) at $W=32$.
The results are presented in Figure~\ref{fig:subsequence_length}.

\textit{Robustness:} As shown in Figure~\ref{fig:subsequence_length}(a), PromptTSS, the strongest stateless baseline, collapses to random guessing ($\sim$35\%) as early as $W=8$ due to the overwhelming number of unprompted windows.
In contrast, Perseus exhibits superior robustness, maintaining high accuracy far longer and yielding a massive performance gap (+57.5pp at $W=8$).
Even at the extreme horizon of $W=32$, Perseus retains useful signal (64.33\%) where the baseline fails completely, demonstrating that the persistent memory bank successfully bridges extended temporal gaps to retrieve user intent.

\textit{Scalability:} Figure~\ref{fig:subsequence_length}(b) reports the empirical runtime trend as subsequence length increases.
Because the total dataset size is fixed, the number of processed timesteps remains approximately constant as the number of subsequences decreases ($M \times L_s \approx \text{const}$).
A model linear in $L_s$ would therefore have roughly constant epoch time, aside from batching and memory interactions.
As $L_s$ increases by $16\times$ (from $W=2$ to $W=32$), training time per epoch increases from 819s to 1310s, which is sublinear in $L_s$ over the evaluated range.
This trend is consistent with avoiding quadratic growth, although it does not establish an asymptotic complexity bound.

\section{Conclusion}
In this work, we introduced Perseus, a framework that transitions interactive time series segmentation from synchronous window-level processing to \textit{asynchronous state management}.
We identified that stateless prompt-based models suffer from \textit{contextual amnesia} when grouped supervision leaves large temporal gaps between user inputs.
Perseus addresses this by decoupling supervision from inference via an asynchronous \textit{Write-Read memory mechanism}.
Specifically, grouped bursts of feedback are asynchronously encoded into a persistent memory bank (Write), which is then actively queried to guide predictions in unprompted windows (Read).
This design creates a semantic bridge that preserves user intent across the entire timeline, ensuring consistent segmentation even when supervision is sparse and distant.

Extensive experiments on six wearable and industrial datasets confirm the value of statefulness under grouped supervision.
Under a strict grouped budget (5\% density), Perseus demonstrates superior robustness, achieving an 85\% accuracy improvement over the evaluated stateless baselines in multi-granularity settings.
Furthermore, in interactive sessions, Perseus exhibits higher data efficiency, yielding an average per-iteration gain of 2.66 percentage points compared to 1.19 for the stateless PromptTSS baseline.
Matched label-propagation analysis further shows that simple state persistence can be competitive on long-duration activity sequences, but fails on industrial datasets with more frequent unprompted transitions.
These results establish that contextual memory, rather than label copying alone, is important for handling realistic, bursty human-in-the-loop interactions.

This study has limitations.
Prompts are simulated from clean annotations, and robustness to noisy or conflicting expert feedback deserves further study.
The current memory bank uses simple accumulation.
Future work can introduce explicit forgetting or conflict resolution.
We also plan to integrate \textit{active learning} to identify informative unprompted windows and reduce the burden on domain experts.

\bibliographystyle{IEEEtran}
\bibliography{references}

\end{document}